# Evaluating the Predictive Performance of Positive-Unlabelled Classifiers: a brief critical review and practical recommendations for improvement

Jack D. Saunders
University of Kent
Canterbury
United Kingdom

Jds39@kent.ac.uk

Alex A. Freitas
University of Kent
Canterbury
United Kingdom

A.A.Fretias@kent.ac.uk

## ABSTRACT

Positive-Unlabelled (PU) learning is a growing area of machine learning that aims to learn classifiers from data consisting of labelled positive and unlabelled instances. Whilst much work has been done proposing methods for PU learning, little has been written on the subject of evaluating these methods. Many popular standard classification metrics cannot be precisely calculated due to the absence of fully labelled data, so alternative approaches must be taken. This short commentary paper critically reviews the main PU learning evaluation approaches and the choice of predictive accuracy measures in 51 articles proposing PU classifiers and provides practical recommendations for improvements in this area.

## 1. INTRODUCTION

Positive-Unlabelled (PU) learning is an area of machine learning that aims to learn classification models from datasets that consist of only positive-class and unlabelled instances, where the latter may be in reality either positive or negative, but their label is unknown [1]. PU learning shares the goal of binary classification – to accurately predict the class of a new instance unseen in the training phase by learning to distinguish between two classes. However, since a standard binary classification algorithm requires a training set with two class labels, a standard classification algorithm learning from a PU dataset would have to treat all unlabelled instances as a separate class. Thus, the learned classification model would predict the probability of an instance being labelled, $\Pr(s = 1)$, as opposed to the probability of an instance belonging to the positive class, $\Pr(y = 1)$ [2], where $s$ is a variable taking the value 1 or 0 to denote whether or not an instance is labelled, and $y$ is the true class label of an instance, taking the value 1 or 0 to denote the positive or negative class respectively. Models learned by a PU learning algorithm, on the other hand, are trained to predict $\Pr(y = 1)$ given PU data.

PU learning is an important area of machine learning as it naturally arises in many different domains, such as bioinformatics [3][4][5], text mining [6][7][8], and cyber security [9][10]. For example, reference [3] utilises PU learning for the prediction of genes associated with diseases. This is a PU learning task where disease-associated genes are positive instances, as confirmed by biomedical experiments. However, the vast majority of the genes not associated with diseases have not undergone such experiments, since these experiments are expensive. As such, the genes not associated with diseases are better thought of as unlabelled instances as there is no experimental evidence indicating either association or disassociation. An example from the domain of text mining is found in [6], which proposed a text classification system utilising PU learning for web page classification. This is another learning task where PU learning is appropriate. Scraping web pages is an easy and quick task, so assembling a large dataset is a simple process. However, the majority of the instances (webpages) will be unlabelled as manually labelling each instance is an expensive task. As illustrated by these examples, PU learning is appropriate whenever the dataset consists of a small sample of reliable positives and a much larger remaining sample of unknown-label instances.

PU learning is related to semi-supervised learning [11] in the sense that it specialises the semi-supervised scenario [1]. In both semi-supervised and PU learning, typically the large majority of training instances is unlabelled; but a semi-supervised learning's training set includes small proportions of both positive and negative instances, whilst a PU learning's training set does not include any negative instance.

Over the past two decades, many PU learning algorithms have been developed for a wide array of applications [1][12][13]. However, little has been written on the subject of evaluation metrics for PU learning, which is a challenging task as discussed next. Hence, in order to address this gap in the literature, this short commentary paper critically reviews the main PU learning evaluation approaches and the choice of predictive accuracy measures in 51 articles using PU learning for the classification task of machine learning, and as a result it provides guidance on how to improve the methodology of evaluation of PU learning methods.

More specifically, this paper provides three main contributions, as follows. First, we discuss the pros and cons of two major approaches for evaluating PU classifiers, namely using genuine PU datasets (with truly unlabelled data) and engineering a PU dataset to create unlabelled data from a standard binary-class dataset (where all instances are labelled); and we note that sometimes predictive accuracy results are reported for genuine PU datasets without mentioning that their validity depends on the validity of a commonly used PU-learning assumption. Second, we argue that, when using the engineered-dataset approach (by far the most popular in the literature), the experiments should consider a wider range of dataset configurations; and to support such more extensive experiments, we provide a GitHub repository of freely available engineered PU datasets. Third, we note that in the reviewed literature on PU learning, the choice of predictive performance measures for reporting results has been in many cases sub-optimal. We identify some measures which should be used more often (explaining why), and also identify a measure whose use should be arguably reduced in the PU-learning literature.

## 2. THE CHALLENGE OF EVALUATING PU LEARNING CLASSIFIERS

The absence of negative instances presents an issue to the evaluation of PU learning models as predictive accuracy metrics usually rely on knowledge of the true class labels of each instance. However, in PU learning we have only knowledge of the true class label of a sample of positive instances. The remaining positive instances, and all negative instances, are unlabelled. Due to these unlabelled instances, popular metrics such as true positive and false negative rates, precision, recall, and the F-measure [14], cannot be correctly calculated.

Before proceeding, it is important to recall the Selected Completely at Random (SCAR) assumption. Formalised by [2], the SCAR assumption states that for the given data, $\Pr(s = 1) = \Pr(s = 1|x)$, where $\Pr(s = 1)$ is the probability of an instance being labelled, and $x$ represents an instance's feature vector. That is, the labelled instances are selected from the positive distribution irrespective of their feature vectors and thus the labelled set is an independent and identically distributed sample from the positive distribution. Or, put simply, the sample of positive instances in the labelled positive set is representative of the entire set of positive instances, both labelled and unlabelled. Under this assumption, we can estimate several performance metrics for models tested on genuine PU data, that is PU data that has not been engineered from a standard positive-negative (PN) dataset (with positive and negative labels). These estimation metrics are discussed in the following section. However, as these metrics represent performance estimates, they are not entirely robust. Arguably, a more robust approach is to evaluate a PU learning method on an engineered PU dataset before applying that method to a genuine PU learning task.

Actually, the approach taken by most of the papers we reviewed (42 out of 51 papers) is to evaluate their proposed method on engineered PU data created from a standard PN dataset by hiding a certain percentage of positive instances in the negative set, thus creating an unlabelled set (i.e., all negatives and the hidden positives will be indistinguishably treated as 'unlabelled'). This is done for the training set, whilst leaving the test set untouched. That is, the test set will contain positive and negative instances as in the original dataset. Hence, the model is trained on PU data but evaluated on fully-labelled data. Therefore, we can accurately calculate standard PN metrics. This is arguably a more robust approach as the performance is not estimated based on the SCAR assumption (i.e., that assumption is not required); we can rely on values of performance metrics that are accurately calculated based on the known class labels of the instances in the test set. However, this approach assumes that a method that demonstrates good performance on an engineered PU dataset will also perform well on a genuine PU dataset.

In practice, in real-world applications, ideally both genuine PU data and engineered datasets should be used, given their different pros and cons, in order to provide a more comprehensive performance evaluation.

## 3. EVALUATING PU LEARNING PERFORMANCE ON VARYING DISTRIBUTIONS OF UNLABELLED POSITIVE INSTANCES

To further analyse the performance of a PU learning algorithm, it is important, when feasible, to assess the performance of its learned model on multiple distributions of unlabelled instances. That is, testing on different versions of the same dataset with differing percentages of the positive instances hidden in the unlabelled set in the training set, when doing experiments with engineered PU datasets. Due to the nature of PU learning, it is often hard to know the distribution of positive instances, and what proportion of them remain unlabelled. However, there are scenarios in which the distribution is known, or can be estimated [15][16]. In such scenarios, by providing results of experiments conducted on multiple distributions, we can provide a more comprehensive analysis of PU methods and inform on their appropriate use case.

Whilst the majority of those papers using engineered datasets use a variety of distributions of unlabelled positive instances, 43% (18/42) use three or less distributions, and 21% (9/42) use a single distribution. Among these 9, there is no consensus of an appropriate distribution by which to assess the proposed method, with the number of positive instances in the training set hidden in the unlabelled set ranging from 15% of positive instances in the case of [6] to all but a single positive instance in the case of [17]. Furthermore, there are a substantial number of papers that, whilst using more than three class distributions, use too narrow a range. For example, those that consider only an extreme case, such as hiding 95-99% of the positive instances in the unlabelled set [18], or hiding only 5%, 10%, 90%, and 95% of positive instances in the unlabelled set [16]. Whilst these values may be appropriate for the use case tested on, performance generalisation is an important factor to consider when proposing a new algorithm. That is, when proposing a PU learning algorithm, ideally researchers should not only report results about their specific task of interest, but also report more comprehensive results which would inform other researchers on the efficacy of the given algorithm in general.

To aid in this, we have provided a GitHub repository consisting of engineered PU datasets with the percentage of unlabelled positive instances varying from 5% to 95% in 5% increments. This repository can be accessed at https://github.com/jds39/Unlabelled-Datasets. To create these datasets, standard PN benchmarking datasets were obtained from various sources (sources are cited for each individual dataset in the repository) and converted to PU datasets by setting the class value of a given proportion of positive instances to "2". These instances are to be treated as unlabelled positive instances in the training set when partitioning data for training and testing. For ease of use, 5 training and test sets have been created for each dataset to implement a stratified 5-fold cross-validation, with the instances with class label "2" set to "0" (negative) in the training set and "1" (positive) in the test set. The repository uses 5 rather than the more usual 10 folds for cross-validation due to the small number of positive instances in some cases, since using 10 folds would cause too few positive instances in each fold.

## 4. A BRIEF REVIEW OF THE MEASURES USED TO EVALUATE PU LEARNING CLASSIFIERS

A review of 51 PU learning papers proposing algorithms from 2002 to 2021 was carried out, identifying the metric by which the proposed algorithm was evaluated. This is summarised in Table 1. Regarding the second column, "E" represents the papers using only engineered PU data, "G" represents papers using only genuine PU data, and "E&G" represents papers using both.

The most commonly reported metric was the F-measure (37 of 51 papers), defined as follows:

$$F - \text{measure} = 2 \times \frac{\text{Precision} + \text{Recall}}{\text{Precision} \times \text{Recall}}$$

$$\text{Precision} = \frac{TP}{TP + FP} \quad \text{Recall} = \frac{TP}{TP + FN}$$

Where TP, FP and FN are the number of true positives, false positives and false negatives, respectively. However, as previously mentioned, the F-measure cannot be precisely calculated when using genuine PU data, because without knowledge of the true class labels in the unlabelled set, TP, FP, and FN cannot be precisely calculated.

When working with genuine PU data, the values of measures like precision, recall and F-measure refer to the performance of a classifier predicting $\Pr(s = 1)$, rather than $\Pr(y = 1)$; hence those values can serve as an estimation of model performance under the SCAR assumption. Elkan and Noto [2] show that a result of the SCAR assumption is Equation (1).

$$f(x) = \frac{g(x)}{c} \quad (1)$$

i.e., $g(x)$ differs from $f(x)$ by a constant factor, where $g(x)$ is a probabilistic classifier trained to distinguish between the labelled and unlabelled sets and thus predicts $\Pr(s = 1|x)$, and $f(x)$ is a probabilistic classifier trained to distinguish between the positive and negative sets and thus predicts $\Pr(y = 1|x)$.

One major implication of Equation (1) is that if we simply aim to rank instances by their predicted probability of belonging to the positive class, e.g., in anomaly detection, we can simply use $g(x)$, as the instances with the highest predicted probability of belonging to the positive class will be the same for both $g(x)$ and $f(x)$. Therefore, under the SCAR assumption, if a model is able to accurately predict the class of the labelled positive instances (high estimated recall) whilst keeping the number of unlabelled instances predicted as positive to a minimum (high estimated precision), it can be assumed that the model is performing well; and that whilst it is predicting $\Pr(s = 1)$, its predictions are differing from $\Pr(y = 1)$ by a constant factor. If the data does not adhere to the SCAR assumption, performance cannot be properly estimated based on this implication.

Of the papers we reviewed, 32 of the 37 that reported F-measure evaluated their model using engineered PU data, as described in Section 2. The 5 papers which reported F-measure but did not use an engineered PU dataset did not note that their results are an estimation of F-measure based on the SCAR assumption as they were using genuine PU data.

F-measure is a good metric for evaluating a PU learning classifier, as PU data tends to have very imbalanced class distributions (with small proportions of positive instances) and F-measure, as the harmonic mean of precision and recall, is not impacted by class imbalance in the same way that accuracy (defined in Equation (2)) is. Few papers, however, report precision and recall separately, as well as F-measure. In fact, of the 51 papers we reviewed, only 5 reported precision, and only 7 reported recall. The importance of reporting these metrics is discussed in the next section.

The second most reported metric was accuracy, calculated as shown in equation 2.

$$\text{Accuracy} = \frac{TP + TN}{TP + FP + TN + FN} \quad (2)$$

Where TN is the number of true negatives. For definitions of TP, FP, and FN, refer to equation 1. Like precision and recall, accuracy cannot be accurately calculated when using genuine PU data, but instead represents an estimation. Accuracy suffers from the very imbalanced nature of most PU learning datasets, where the labelled positive set is typically a very small proportion of the total number of instances. As such, a high accuracy can be trivially achieved simply by classifying all instances as negative [58]. Hence, the popularity of this measure in PU learning is surprising, and its use should be reduced.

The third most reported metric was the area under the Receiving Operator Characteristic curve (AUROC). The ROC curve shows the true positive rate (TPR) plotted against the false positive rate (FPR) for varying classification thresholds between 0 and 1.

$$\text{TPR} = \frac{TP}{TP + FN} \quad \quad \text{FPR} = \frac{FP}{FP + TN}$$

As with precision and recall, the values comprising the ROC curve are estimations; and the AUROC, when calculated on genuine PU data, is therefore an estimation based on the SCAR assumption.

## 5. PRIORITISATION VS ANOMALY DETECTION

There are two primary goals of PU learning - prioritisation and anomaly detection. Depending on the goal, either precision or recall may be more important than the other. As explained in [24], if the goal of the learning task is prioritisation, precision is the most important metric. If the goal is anomaly detection, recall is the most important metric, as follows.

In the task of prioritisation, one wishes to identify highly ranked targets. That is, instances that have the highest predicted probability of belonging to the positive class. As we are interested in prioritising instances, it is important that our model identify few unlabelled positives. Prioritisation is required when we need to identify a few top-ranked (most likely positive) instances for performing expensive or time-consuming future experiments on those few high-priority instances, and so minimising the number of false positives (maximising precision) is particularly important, to avoid doing experiments that produce negative results. Maximising recall is not so crucial because it would be too expensive or too time-consuming to perform future experiments to validate a large number of instances predicted as positives. An example is gene prioritization, where each gene is an instance and the positive class represents a biological function (or associated disease) of the gene, since biological experiments to verify gene functions tend to be expensive and time-consuming.

| Reference | Engineered or Genuine PU Data | F-measure | Accuracy | Precision | Recall | AUROC |
|---|---|---|---|---|---|---|
| [19] | E | ✓ | · | · | · | · |
| [20] | E | ✓ | · | · | · | · |
| [21] | E | ✓ | ✓ | · | · | · |
| [22] | E | · | ✓ | · | · | · |
| [23] | E | ✓ | · | · | · | · |
| [24] | E | · | · | · | · | · |
| [25] | E | · | ✓ | · | · | · |
| [16] | E | · | ✓ | · | · | · |
| [2] | E | ✓ | · | · | · | ✓ |
| [26] | E | ✓ | · | · | · | · |
| [27] | E | ✓ | · | · | · | · |
| [28] | E | · | ✓ | · | · | · |
| [29] | E | ✓ | ✓ | · | · | · |
| [30] | E | · | ✓ | · | · | · |
| [31] | E | ✓ | · | · | · | · |
| [32] | E | · | ✓ | ✓ | ✓ | · |
| [33] | E | ✓ | ✓ | · | · | · |
| [7] | E | ✓ | · | · | · | · |
| [34] | E | ✓ | · | · | · | · |
| [35] | G | · | · | · | · | ✓ |
| [36] | E | ✓ | ✓ | · | · | · |
| [37] | E | ✓ | · | · | · | · |
| [38] | G | · | ✓ | · | · | · |
| [39] | E | ✓ | ✓ | · | · | · |
| [40] | E | ✓ | · | · | · | · |
| [41] | E | ✓ | · | · | · | · |
| [42] | E | ✓ | · | · | · | · |
| [43] | E&G | ✓ | ✓ | · | · | ✓ |
| [44] | E | ✓ | ✓ | · | · | · |
| [45] | E | ✓ | · | · | · | · |
| [8] | E | ✓ | · | · | · | · |
| [6] | E | · | ✓ | · | · | ✓ |
| [10] | G | ✓ | ✓ | · | ✓ | · |
| [46] | E&G | · | · | · | · | ✓ |
| [17] | E | ✓ | · | · | · | · |
| [3] | G | ✓ | · | ✓ | ✓ | · |
| [47] | E | ✓ | · | · | · | · |
| [48] | E | ✓ | ✓ | · | · | · |
| [4] | G | ✓ | · | ✓ | ✓ | ✓ |
| [49] | E | ✓ | ✓ | · | · | · |
| [5] | G | ✓ | · | · | · | · |
| [50] | E&G | ✓ | · | · | ✓ | · |
| [51] | E | ✓ | · | · | · | · |
| [52] | G | · | · | ✓ | ✓ | · |
| [53] | E | ✓ | · | · | · | · |
| [54] | E | ✓ | · | · | · | · |
| [9] | G | · | ✓ | · | · | · |
| [55] | E | ✓ | · | · | · | · |
| [56] | G | ✓ | · | ✓ | ✓ | ✓ |
| [18] | E | ✓ | · | · | · | · |
| [57] | E | · | · | · | · | ✓ |
| **Totals** | E:42 G:12 | 37 | 19 | 5 | 7 | 8 |

**Table 1: Evaluation metrics used by papers proposing PU learning algorithms**

In the task of anomaly detection, one wishes to accurately identify positive class instances, which are usually a very small minority ("anomalies"). In anomaly detection, usually the cost of a false negative is usually much higher than the cost of a false positive. Therefore, maximising recall (minimising false negatives) is usually more important than maximising precision (minimising false positives). For example, when classifying a bank's transactions into fraud (anomaly) vs non-fraud (normal), the cost of misclassifying a fraud transaction as a non-fraud transaction is usually much higher than vice-versa.

Table 2 shows, for the 12 papers from Table 1 using real-world datasets for PU learning (i.e., genuine PU datasets, rather than engineered PU datasets), whether their goal is anomaly detection or prioritisation and whether they reported precision or recall. Unfortunately, out of the 3 papers addressing anomaly detection in Table 2, only one is reporting recall, and out of the 9 papers addressing prioritisation, only 4 are reporting precision. Without these results, the suitability of the proposed method for the given target application cannot be determined. This shows that the importance of reporting precision and recall separately (particularly in prioritisation or anomaly detection tasks) is still not well appreciated in the PU learning area.

Whilst precision and recall may be important metrics for a given learning task, considering either precision or recall in isolation is flawed, since it is well-known that it is relatively easy to maximise one of these measures at the expenses of obtaining a poor value for the other. Hence, it is important to report the F-measure, precision, and recall. This will allow researchers looking to utilise a PU learning method to make an informed decision on which algorithm is most appropriate for their use case, favouring those with a high F-measure and precision for prioritisation tasks, and those with a high F-measure and recall for anomaly detection.

| Reference | Anomaly detection | Prioritisation | Precision | Recall |
|---|---|---|---|---|
| [35] | . | ✓ | . | . |
| [38] | ✓ | . | . | . |
| [43] | . | ✓ | . | . |
| [10] | ✓ | . | . | ✓ |
| [46] | . | ✓ | . | . |
| [3] | . | ✓ | ✓ | ✓ |
| [4] | . | ✓ | ✓ | ✓ |
| [5] | . | ✓ | . | . |
| [50] | . | ✓ | . | ✓ |
| [52] | . | ✓ | ✓ | ✓ |
| [9] | ✓ | . | . | . |
| [56] | . | ✓ | ✓ | ✓ |
| **Total** | 3 | 9 | 4 | 6 |

**Table 2: PU learning goals in reviewed papers using genuine PU data**

In real-world applications, the "ideal" approach to PU learning model evaluation would be to train the model on genuine PU data, and then verify the performance of the model by learning the true labels of the unlabelled instances, or at least a subset of unlabelled instances which were predicted with a high probability to be positive instances. However, this is unfeasibly expensive for the vast majority of PU learning tasks. If it were not, there would be no need for learning a PU classifier in the first place as one could simply label the unlabelled data and use a traditional classifier. Therefore, the two approaches identified, estimating performance using genuine PU data and experimenting on engineered PU datasets, offer more feasible approaches to reliable model evaluation and provide a greatly needed framework for the evaluation of PU learning classifiers.

## 6. CONCLUSIONS

This paper has reviewed evaluation approaches for PU learning classifiers. We have found that the majority of reviewed papers utilised an evaluation approach whereby the classifier is evaluated on an engineered PU dataset, where some positive instances are 'hidden' as unlabelled instances in the training set but not in the test set, thus allowing for accurate calculation of standard evaluation metrics as the class labels of the instances in the test set are known. However, we noted that many studies consider a single or a few possible percentages of positive instances 'hidden' as unlabelled instances in the training set. We argue that, for a more comprehensive analysis of PU classifiers, a wider range of that percentage should be used in the experiments; and to support such experiments, we provide a GitHub repository of freely available engineered PU datasets. In addition, when using the alternative approach of evaluating PU classifiers on genuine PU datasets (with originally unlabelled data), we noted that sometimes predictive accuracy results are reported without mentioning that those results' validity depends on the SCAR assumption, an important assumption in PU learning.

Regarding the choice of predictive accuracy measures, we found that the most used measure in the reviewed papers was the F-measure, which is indeed appropriate for PU learning given the large class imbalance that typically occurs in PU datasets. Unfortunately, however, there was overall a significant lack of reporting precision and recall separately. We argue that both precision and recall should be reported independently (in addition to the F-measure) to allow for readers to make an informed decision regarding the appropriate use case for a classifier. As described, classifiers that demonstrate a high recall should be considered for anomaly detection tasks, whilst classifiers that demonstrate a high precision should be considered for prioritisation tasks.

## About the authors:

**Jack D. Saunders** is a PhD student at the University of Kent, supervised by Prof. Alex. Freitas. Jack Saunders is supported by an EPSRC-funded PhD studentship.

**Prof. Alex A. Freitas** has an interdisciplinary background: a PhD in Computer Science (University of Essex, UK, 1997) and a master's degree (MPhil) in Biological Sciences (University of Liverpool, UK, 2011